\pdfoutput=1

\documentclass[11pt]{article}

\usepackage[]{acl}

\usepackage{times}
\usepackage{latexsym}

\usepackage[T1]{fontenc}

\usepackage[utf8]{inputenc}

\usepackage{microtype}

\usepackage{inconsolata}

\usepackage{algorithm}
\usepackage{algpseudocode}
\usepackage{amsmath}
\usepackage{pgf}
\newcommand{\timeformat}[1]{%
    \pgfmathtruncatemacro\minutes{#1/60}%
    \pgfmathtruncatemacro\seconds{#1-\minutes*60}%
    \ifnum\seconds<10 \def\secondsdigit{0}\else\def\secondsdigit{}\fi
    \minutes:\secondsdigit\pgfmathprintnumber{\seconds}%
}

\usepackage{fp}
\newcommand{\speedupSGD}[1]{%
    \FPeval{\result}{clip(1335.6/#1)}
    \FPround{\result}{\result}{1}
    x\result %
}

\usepackage{fp}
\newcommand{\speedupDART}[1]{%
    \FPeval{\result}{clip(1990.4/#1)}
    \FPround{\result}{\result}{1}
    x\result %
}

\title{Control-DAG: Constrained Decoding for Non-Autoregressive Directed Acyclic T5 using Weighted Finite State Automata}

\author{Jinghong Chen, Weizhe Lin, Jingbiao Mei, Bill Byrne \\
        Department of Engineering \\ University of Cambridge \\ \texttt{\{jc2124, wl356, jm2245, wjb31\}@cam.ac.uk}}

\begin{document}
\maketitle
\begin{abstract}

The Directed Acyclic Transformer is a fast non-autoregressive (NAR) model that performs well in Neural Machine Translation. Two issues prevent its application to general Natural Language Generation (NLG) tasks: frequent Out-Of-Vocabulary (OOV) errors and the inability to faithfully generate entity names. We introduce Control-DAG, a constrained decoding algorithm for our Directed Acyclic T5 (DA-T5) model which offers lexical, vocabulary and length control. We show that Control-DAG significantly enhances DA-T5 on the Schema Guided Dialogue and the DART datasets, establishing strong NAR results for Task-Oriented Dialogue and Data-to-Text NLG.

\end{abstract}

\section{Introduction}

Non-autoregressive (NAR) models for text generation offer the promise of much faster generation  than auto-regressive (AR) models.  However NAR models have been largely developed for Neural Machine Translation (NMT)~\cite{NARTextGenerationSurvey_2022}, with other Natural Language Generation (NLG) tasks less well studied. We will show how a NAR model developed for NMT, the Directed Acyclic Transformer (DAT)~\citep{DA-Transformer-pmlr-v162-huang22m}, can be used for generation in Task-Oriented Dialogue (TOD) and Data-to-Text (D2T) scenarios. 

DATs as originally developed for NMT perform poorly in NLG on TOD and D2T tasks: they fail to generate specified entity names in up to 40\% of responses and frequently (>20\%) produce Out-Of-Vocabulary (OOV) words. Practical systems must operate at zero error rate in these aspects to be deployable at scale. Previous NAR study reported similar error patterns~\cite{NARTextGenerationSurvey_2022}. Unless these shortcomings are addressed, NAR models will not be usable for general NLG.



We introduce three constrained decoding procedures for NLG using DATs.   Our approach converts Directed Acyclic Graphs (DAG) generated by DAT into Weighted Finite State Automata (WFSA).  We then intersect these WFSAs with other automata that are defined to ensure that designated entities (\textit{lexical constraints}) are generated and OOVs are eliminated (\textit{vocabulary constraints}). To avoid generating responses that are too short, we employ a Viterbi decoding algorithm to control the target length of the generated text (\textit{length constraints}).   

We refer to the decoding procedure that incorporates all these steps as  \texttt{Control-DAG}.  We evaluate extensively on the Schema Guided Dialogue (SGD)~\citep{SGD-dataset-rastogi2020} and the Data Record To Text (DART) datasets~\citep{DART_dataset_paper} for NLG in TOD and D2T domains. Our Directed Acyclic T5 model, when decoded with \texttt{Control-DAG}, is free from OOV error, faithfully generates all specified entity names, and achieves marked BLEU and BLEURT gains on both datasets. 
We use \texttt{pynini}~\cite{pynini_2016} for WFSA operations. 
Our contributions are summarized below:
\begin{enumerate}
    \item We introduce \texttt{Control-DAG}, a constrained decoding algorithm which simultaneously offers lexical, vocabulary, and length controls for Directed Acyclic models, addressing key limitations in NAR text generation. 
    \item We demonstrate the effectiveness of \texttt{Control-DAG} on two major NLG tasks: Task-Oriented Dialogues and Data-to-Text. To our knowledge, DA-T5 with \texttt{Control-DAG} is the first practical NAR benchmark on the SGD and the DART datasets.\footnote{Code: \href{https://github.com/EriChen0615/ControlDAG}{https://github.com/EriChen0615/ControlDAG}}
\end{enumerate}

\begin{figure*}[h!]
    \centering
\includegraphics[width=1.05\textwidth]{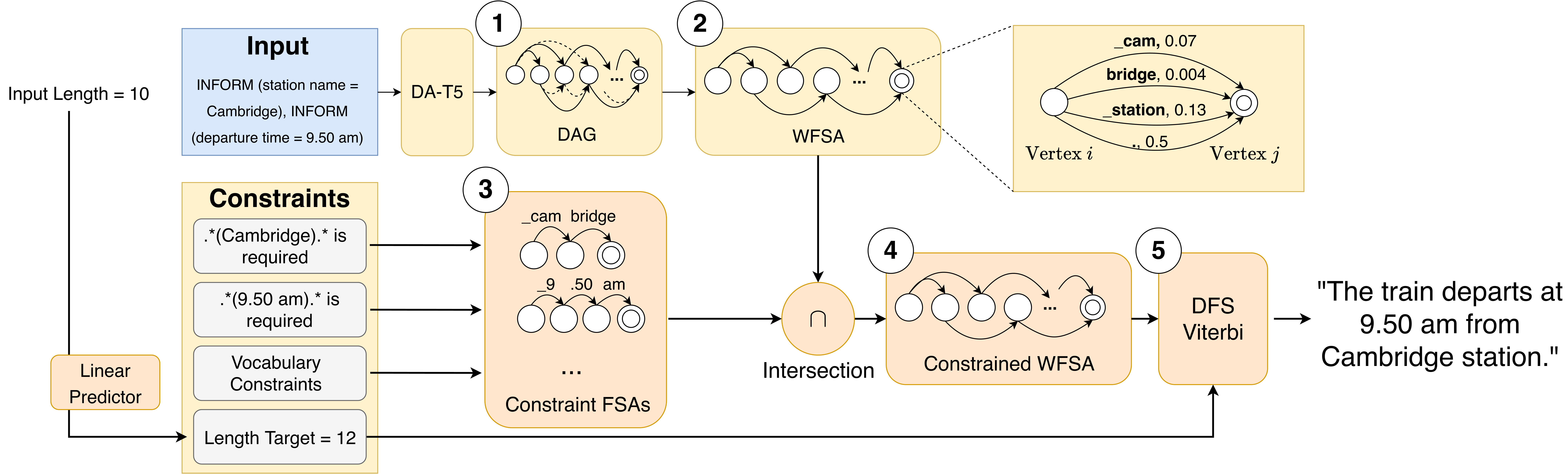}
    \caption{Control-DAG with lexical, vocabulary, and length constraints. \textbf{1.} Directed Acyclic T5 (DA-T5) takes the input text to generate a Directed Acyclic Graph (DAG). \textbf{2.} The DAG is pruned by likelihood, keeping $K_e$ most likely output tokens and $K_t$ most likely out-going arcs, and converted into a Weighted Finite State Automaton (WFSA). We show WFSA vertices and arcs in the upper-right corner. \textbf{3.} For lexical and vocabulary constraints,  constraint FSAs are built from equivalent regular expressions (Sec.3.1). The length target predictor is a simple linear predictor based on the input sequence length (Sec.4).  \textbf{4.} We intersect the WFSA with constraint FSAs to obtain a constrained WFSA which only contains hypotheses that satisfy all lexical and vocabulary constraints. \textbf{5.} DFS-Viterbi is used to obtain the most likely string in the constrained WFSA that satisfies the length constraint.}
    \label{fig:main-diagram}
\end{figure*}

\section{Related Work}

The Directed Acyclic Transformer (DAT)~\cite{DA-Transformer-pmlr-v162-huang22m} performs on par with AR baselines in NMT and has attracted much interests. \citet{Viterbi_decoding_for_Da-Transformer} developed a Viterbi decoding algorithm for DAT. \citet{FuzzyAlignmentsInDATransformerMa2023} introduced a fuzzy alignment objective to improve DAT training. In NLG,  PreDAT~\citep{PreDAT_pretraining_da_transformer_huang_2023} pretrains a DAT for open-domain dialogue, notably with high word error rate reported even after extensive pre-training. Our work highlights the links between DATs and automata, and shows well-studied WFSA algorithms~\cite{MohriWFST_SpeechRecognition_2002} can be used in constrained decoding to eliminate OOV errors. 

Enforcing lexical constraints in auto-regressive decoding has been studied extensively.  {\em Constrained beam search} (CBS)~\cite{DynamicBeamAllocation(DBA)_2018, VectorizedDBA_2019, DisjuctivePositiveConstraints(DPC)_2020} is a widely used family of lexically constrained decoding procedure. We show how CBS can be adapted to NAR Directed Acyclic models.  


\section{Constrained Decoding with DA-T5}

 The architecture of our \textit{DA-T5} model follows that of the DAT by \citet{DA-Transformer-pmlr-v162-huang22m}. Conceptually, DAT takes an input sequence and generates a DAG with a pre-determined number of DAG vertices. Vertex embeddings are produced first, and then token emission probabilities and state transition probabilities are  generated from these vertex embeddings via softmax and self-attention, resp.  Each vertex has a token emission distribution. These vertices and transitions define a weighted DAG that contains output string hypotheses. DAT uses a vanilla Transformer to produce vertex embeddings whereas we use T5, hence the name DA-T5. 
 
 In training DA-T5,  we use  `glancing training' \cite{GlancingTransformer_GLAT_2021} as DAT. In inference, DAGs are generated with DA-T5 and converted to WFSAs. The procedure is simply Moore-to-Mealy Machine conversion (Appendix~\ref{app:dag-to-wfsa_algo_detail}). Prior to the conversion, we perform likelihood-based pruning of each vertex, keeping $K_e$ most likely output tokens and $K_t$ most likely out-going arcs. This pruning balances coverage against decoding speed, with larger thresholds leading to a more complete WFSA at the cost of slower decoding. 

\subsection{Constrained Decoding}
\label{sec:method_constrained_decoding}

\begin{table*}[!htp]
\centering
\vspace{-0.3cm}
\begin{tabular}{cl|ccccccc}
\hline 
\textbf{\#} & \textbf{Decoding} & \textbf{BLEURT} & \textbf{BLEU} & \textbf{BLEU-BP} & \textbf{NEO}$\downarrow$ & \textbf{SER}$\downarrow$ & \textbf{Time} & \textbf{Spd. Up} \\ \hline
\hline \multicolumn{9}{c}{\emph{T5-small (Auto-regressive)}} \\ \hline
1 & Greedy & 69.7 & 28.8 & 1.00 & \textbf{0.0} & 0.49 & \timeformat{810.2} & \speedupSGD{810.2} \\
2 & Beam search (BS) & \textbf{70.2} & \textbf{29.1} & 1.00 & \textbf{0.0} & 0.12 & \timeformat{965.0} & \speedupSGD{965.0} \\
3 & Constrained beam (CBS) & 65.6 & 22.5 & 1.00 & \textbf{0.0} & \textbf{0.0} & \timeformat{1335.6} & \speedupSGD{1335.6} \\ \hline
\multicolumn{9}{c}{\emph{Directed Acyclic T5-small (Non-Autoregressive)}} \\ \hline
4 & Greedy & 56.0 & 18.3 & 0.92 & 29.7 & 46.3 & \timeformat{172.1} & \speedupSGD{172.1} \\
5 & Beam search & 55.6 & 16.0 & 0.60 & 20.7 & 20.6 & \timeformat{410.6} & \speedupSGD{410.6} \\
6 & CBS-DAG & 59.8 & 21.7 & 0.73 & 19.2 & \textbf{0.0} & \timeformat{357.5} & \speedupSGD{357.5} \\ \hline
7 & WFSA shortest path & 53.8 & 13.0 & 0.44 & 12.2 & 34.8 & \timeformat{184.1} & \speedupSGD{184.1} \\
8 & ~~\textit{w/ HLC} & 58.1 & 20.2 & 0.58 & 11.0 & \textbf{0.0} & \timeformat{316.2} & \speedupSGD{316.2} \\
9 & ~~\textit{w/ VC} & 54.0 & 14.1 & 0.45 & \textbf{0.0} & 47.5 & \timeformat{258.8} & \speedupSGD{258.8} \\
10 & ~~\textit{w/ LC (DFS-Viterbi)} & 58.5 & 20.8 & 1.00 & 21.9 & 45.8 & \timeformat{211.7} & \speedupSGD{211.7} \\
11 & Control-DAG & 60.0 & 22.9 & 1.00 & \textbf{0.0} & \textbf{0.0} & \timeformat{794.9} & \speedupSGD{794.9} \\
\hline
\end{tabular}
\caption{\label{tab:sgd-main-results}
Main results on the SGD dataset. For reference, auto-regressive T5-small by \citet{T2G2_TODS_Kale_and_Rastogi} achieves 26.2 BLEU and 0.80 SER. \textbf{BP} stands for the brevity penalty term in computing BLEU. \textbf{SER} stands for Slot Error Rate in percentage. All speed ups are computed against auto-regressive constrained beam search. \textbf{Constrained beam search} (Row 3) forces the replication of slot values that need to appear exactly and hence has zero slot error rate. \textbf{CBS-DAG} (Row 6) refers to Constrained beam search adapted for Directed Acyclic Graph introduced in Sec.\ref{sec:method_constrained_decoding}. \textbf{HLC} refers to Hard Lexical Constraint; \textbf{VC} is Vocabulary Constraint; and \textbf{LC} is Length Constraint. \textbf{Control-DAG} (Row 11) is WFSA shortest path decoding with HLC, VC, and LC applied simultaneously. 
}
\end{table*}


For hard lexical and vocabulary constraints we build corresponding Finite State Automata (FSA). Intersecting the WFSA with these constraint FSAs produces a WFSA that only contains hypotheses that satisfy all constraints~\cite{MohriWFST_SpeechRecognition_2002}. 
For length constraints, we propose a pruned version of DAT Viterbi decoding by \citet{Viterbi_decoding_for_Da-Transformer} to search for strings with specified length. Appendix \ref{app:all_algo_details} gives implementation details and complexity analyses. Figure \ref{fig:main-diagram} illustrates our Control-DAG system with an example.

\paragraph{Hard Lexical Constraints (HLC)}
For each phrase $C_i$ that must appear in the generation, we construct a constraint FSA $A_i$ that accepts and only accepts strings where the phrase $C_i$ appears at least once, corresponding to the regular expression ``$.\ast(C_i).\ast$''~\cite{RegularExpressionIEEEStandard}. 
We then intersect the WFSA converted from the DAG with all of the constraint FSAs. The resulting WFSA $W_{HLC}$ contains only hypotheses that satisfy all lexical constraints. 


\paragraph{Vocabulary Constraints (VC)} We build a vocabulary FSA $A_{vocab}$ that accepts and only accepts strings of words from a valid vocabulary; intersection with $A_{vocab}$ prevents OOV errors.  $A_{vocab}$ is obtained from three FSAs: a dictionary FSA $A_{dict}$ that accepts and only accepts English words; a special token FSA $A_{spec}$ that accepts and only accepts numbers, punctuation, and special tokens; and a dynamic FSA $A_{dyn}$ that accepts and only accepts entity names specified in the input. The final vocabulary FSA $A_{vocab}$ is obtained by unioning the three FSAs and taking the Kleene closure (Eq.\ref{eq:union for vocabulary FSA}).

\begin{equation}
    A_{vocab} = (A_{dict} \cup A_{spec} \cup A_{dyn})^*
    \label{eq:union for vocabulary FSA}
\end{equation}

For efficiency, we perform a one-time determinization and minimization~\cite{MohriWFST_SpeechRecognition_2002} of the union ($A_{dict} \cup A_{spec}$) and store the optimized FSA in memory. 

\paragraph{Length Constraints (LC)} \citet{Viterbi_decoding_for_Da-Transformer} introduced a Viterbi decoding procedure for DAT that finds the highest scoring hypothesis for each string length.  We find this exact Viterbi procedure to be impractical because the number of WFSA states can be large  (>30,000) after intersection with the constraint  FSAs.  We introduce a pruned version of this procedure, \emph{Depth-First Search Viterbi (DFS-Viterbi)}. DFS-Viterbi searches the WFSA with DFS and keeps the best hypotheses of all possible string lengths at each vertex to avoid repeated computation. During DFS, we only explore the minimal set of out-going edges such that their cumulative probability is bigger than a threshold $p$. This pruning is inadmissible but works well in practice. We also introduce an exponential length penalty that penalizes strings shorter than target length $L_{tgt}$ and select the hypothesis with the lowest overall costs.  In experiments to follow, $L_{tgt}$ is obtained via simple linear regression. 

\paragraph{HLC with CBS} In addition to automata-based methods, we introduce CBS-DAG, a constrained beam search algorithm for our NAR DA-T5. CBS-DAG is straight-forwardly adapted from AR CBS by \citet{VectorizedDBA_2019} (Appendix \ref{app:CBS-DAG_details}).



\section{Experiments and Results}

We evaluate on the SGD and the DART datasets. 
In SGD, the aim is to generate natural utterances from dialogue actions (e.g.,  \texttt{INFORM(destination=Cambridge)}) that contain the specified information. DART is a more general data-to-text task that takes triplets of \texttt{(SUBJECT, RELATION, OBJECT)} to generate natural texts. Hyper-parameters and implementation details are in Appendix \ref{app:experiment_setup_details}. 


\paragraph{Metrics}  We use BLEURT~\cite{BLEURT_paper} and BLEU~\cite{BLEU_metric_paper} to measure text quality relative to ground truth text. We also report the BLEU \emph{Brevity Penalty (BP)}, as a small BP indicates too short generation. 
For SGD, we use \textit{Slot Error Rate (SER)}~\cite{T2G2_TODS_Kale_and_Rastogi} to evaluate lexical faithfulness. A slot error occurs when a slot value that should be reproduced exactly (e.g., a phone number) is not in the generated text. 
For DART, we use subjects/objects whose string values are always in the ground-truth training text as hard lexical constraints and propose \textit{Exact Occurrence error Rate (EOR)} for evaluation. EOR is the percentage of model responses where at least one of the string values from these subjects/objects is missing. 
For OOV errors, we define \emph{neologism rate (NEO)} to be the percentage of model's responses that contain at least one OOV generation. 

We emphasize that SER, EOR, and OOV are critical metrics as even a small error rate could lead to an intolerable number of misleading responses for systems deployed at scale. 
`Speed up' is measured against auto-regressive CBS implemented by \citet{DisjuctivePositiveConstraints(DPC)_2020} with batch size of 1 to reflect a realistic NLG system that operates at zero SER/EOR.

\paragraph{Training} We train DA-T5 from scratch by glancing training by~\citet{GlancingTransformer_GLAT_2021} on the SGD and the DART datasets for 30 and 50 epochs, respectively. Auto-regressive T5 is trained following \citet{SchemeGuidedSemanticAccuracy_Chen_2023}. 

\paragraph{Decoding configurations} We use $K_t=K_e=3$ and $K_t=K_e=5$ for DAG-to-WFSA conversion on SGD and DART, respectively. For LC, we fit a simple linear regression model on the training set to predict the target token length given the input token length. 
Decoding hyper-parameters are determined on the validation sets.

\subsection{Non-Autoregressive NLG on SGD}
\label{sec:Results on Schema Guided Dialogue}

\begin{table}[!htbp]
\centering
\hspace*{-0.4cm}
\begin{tabular}{l|ccccc}
\hline
Decoding & BLEURT & BLEU & NEO & SER \\ \hline
Greedy & 56.0 & 18.3 & 29.7 & 46.3 \\
Lookahead & 56.6 & 19.3 & 23.0& 44.6\\
Viterbi & 52.7 &  13.4 & 12.4& 50.5\\ 
Joint Viterbi& 52.1& 12.6 & 10.5& 50.6\\ \hline
Control-DAG & \textbf{60.0} & \textbf{22.9} & \textbf{0.00} & \textbf{0.00} \\ \hline
\end{tabular}
\caption{Performance on the SGD dataset using Control-DAG and other decoding algorithms in the literature. NEO stands for Neologism rate.  \citet{DA-Transformer-pmlr-v162-huang22m} proposed Lookahead. \citet{Viterbi_decoding_for_Da-Transformer} introduced Viterbi and Joint Viterbi.}
\label{tab:compare CONTROL-DAG with other decodings}
\end{table}


Table~\ref{tab:sgd-main-results} reports NLG performance on SGD with auto-regressive T5 decoding in Rows 1-2 with greedy and beam search.  Although these systems yield high BLEURT and BLEU,  they still 
commit slot errors (SER=0.12\%). 
Constrained Beam Search (CBS) eliminates slot errors by forcing the generation of designated slot values, but with longer decoding times (16:05 $\rightarrow$ 22:15) and a degradation in BLEU ($-6.6$) and BLEURT ($-4.6$) compared to unconstrained beam search. This constraint-quality trade-off is also observed in previous study~\cite{DynamicBeamAllocation(DBA)_2018}; See Appendix \ref{app:qualitative_study} for CBS failure modes. Auto-regressive T5 is completely free from OOV errors (NEO=0.0).

Turning to non-autogressive NLG, generation with DA-T5 using common decoding methods (greedy, beam search) leads to very high SER (> 20\%) and OOV errors in at least 20\% of the generated responses (Rows 4, 5). Although our CBS-DAG (Row 6) eliminates SER by design and enhances quality as measured by BLEURT (+3.8) and BLEU (+3.4), its neologism rate is still unusably high (19.2\%). 

We now discuss the performance of our constrained decoding methods. 
Unconstrained WFSA shortest path decoding (Row 7) is as fast as greedy decoding, showing that DAGs can be efficiently converted to WFSAs. However, unconstrained generation directly from the WFSA frequently leads to 
slot errors (SER=34.8\%), OOV errors (NEO=12.2\%), and a harsh brevity penalty (BP=0.44). 
These aspects of text quality can be improved individually by constrained decoding (Rows 8-10): Hard Lexical Constrained decoding eliminates slot errors (SER=0); Vocabulary constraints  eliminate OOV errors (NEO=0); and Length constrained decoding leads to better text lengths (BP=1.0).
\texttt{Control-DAG} (Row 11) combines these methods to  achieves zero SER and zero neologism rate while satisfying the length requirement and yielding a speed advantage of x1.7 relative to auto-regressive CBS.


Table~\ref{tab:compare CONTROL-DAG with other decodings} shows the performance of using existing decoding procedures developed for DA-Transformer to decode DA-T5 on the SGD dataset.  \texttt{Control-DAG} 
has the overall best BLEU (22.9) and BLEURT (60.0) . 

\subsection{Results on DART}
\label{sec:Results on DART}


\begin{table*}[htbp!]
\centering
\begin{tabular}{cl|ccccccc}
\hline 
\textbf{\#} & \textbf{Model} & \textbf{BLEURT} & \textbf{BLEU} &  \textbf{BP} & \textbf{NEO}$\downarrow$ & \textbf{EOR}$\downarrow$ & \textbf{Time} & \textbf{Spd. Up} \\
\hline 
\multicolumn{9}{c}{\emph{T5-small (Auto-regressive)}} \\ \hline
1 & Greedy & 71.2 & 31.3 & 0.95 & 4.1 & 5.0 & \timeformat{1490.2} & \speedupDART{1490.2} \\
2 & Beam search & 72.8 & 31.9 & 0.93 & 3.2 & 3.9 & \timeformat{1853.4} & \speedupDART{1853.4} \\
3 & Constrained beam & 70.5 & 29.3 & 0.95 & 3.3 & \textbf{0.0} & \timeformat{1990.4} & \speedupDART{1990.4} \\ \hline
\multicolumn{9}{c}{\emph{Directed Acyclic T5-small (Non-Autoregressive)}} \\ \hline
4 & Greedy & 45.0  & 18.2 & 1.00 & 48.9 & 39.5 & \timeformat{197.1} & \speedupDART{197.1} \\
5 & Beam search & 45.6 & 14.0 & 0.53 & 34.3 & 43.6 & \timeformat{569.4} & \speedupDART{569.4}\\
6 & CBS-DAG & 46.0 & 18.9  & 0.80 & 36.1 & 0.0 & \timeformat{446.5} & \speedupDART{446.5} \\ \hline
7 & WFSA shortest & 42.1 & 10.8 & 0.38 & 27.3 & 45.4 & \timeformat{229.8} & \speedupDART{229.8} \\
8 & ~~\textit{w/ HLC} & 46.8 & 14.4 & 0.46 & 24.4 & \textbf{0.0} & \timeformat{579.7} & \speedupDART{579.7} \\
9 & ~~\textit{w/ VC} & 39.3 & 7.7 & 0.28 & \textbf{0.0} & 45.1 & \timeformat{638.9} & \speedupDART{638.9} \\
10 & ~~\textit{w/ LC (DFS-Viterbi)} & 46.8 & 18.3 & 0.86 & 44.4 & 40.3 & \timeformat{326.5} & \speedupDART{326.5} \\
11 & CONTROL-DAG & \textbf{46.8} & \textbf{19.0} & 1.00 & \textbf{0.0} & \textbf{0.0} & \timeformat{1443.2} & \speedupDART{1443.2} \\
\hline
\end{tabular}
\caption{\label{tab:DART-main-results-app}
Results on the DART dataset. The naming convention for metrics and decoding methods follow that in Table \ref{tab:sgd-main-results}. \textbf{EOR} is Exact Occurrence Error.
}
\end{table*}

The results on DART (Table \ref{tab:DART-main-results-app}) validate our findings on the SGD dataset: \texttt{Control-DAG} yields the best performance while maintaining a speed advantage and 
each constrained decoding step contributes as expected. We now contrast performance on DART and SGD to show how \texttt{Control-DAG} performs on tasks with very different characteristics.

DART has a challenging vocabulary that causes even AR models to commit OOV errors. This is also reflected by the much higher neologism rate when decoding DA-T5 with greedy (48.9\% versus 29.7\% in SGD). This explains why less aggressive pruning (top-5) is needed for DART relative to SGD (top-3). We find the simple procedure of searching the training data for subjects/objects whose values are exactly reproduced and using them as lexical constraints boosts DA-T5 performance by +4.7 BLEURT and +3.6 BLEU (Row 8, Table \ref{tab:DART-main-results-app}). This demonstrates that hard lexical constraints are effective and easy to apply for less lexically constrained NLG tasks such as DART.

\section{Conclusion}

We propose \texttt{Control-DAG} for decoding non-autoregressive Directed Acyclic models with lexical, vocabulary, and length constraints, addressing key limitations in NAR text generation. Constrained decoding is efficiently performed via well-studied Weighted Finite State Automata algorithms. DA-T5 with \texttt{Control-DAG} establishes strong NAR results on the Schema Guided Dialogue and the DART datasets, bridging gaps in NAR research.

\section{Acknowledgement}
Jinghong Chen is supported by the Warwick Postgraduate Studentship from Christ's College and the Huawei Hisilicon Studentship for the undertaking of the PhD in Engineering at the University of Cambridge.

Weizhe Lin was supported by a Research Studentship funded by Toyota Motor Europe (RG92562(24020)). 

Prof. Bill Byrne holds concurrent appointments as a Professor of Information Engineering at Cambridge University and as an Amazon Scholar.  This publication describes work performed at Cambridge University and is not associated with Amazon.

We would also like to thank all the reviewers for their knowledgeable reviews.

\section{Limitation}

Given our focus on decoding algorithms, we leave further training and model scaling to future work. It is possible to further improve inference speed by writing the DAG-to-WFSA conversion and the DFS-Viterbi algorithm in the C programming language to reduce overhead from the python interface. In this paper, we demonstrate substantial speed-up can be achieved without these optimizations and leaves further speed-up techniques to future work. 

\newpage

\section{Ethical Statement}

We trained two versions of the DA-T5 model: one on the training set of Schema Guided Dialogue and one on the training set of the DART dataset. These are English datasets and do not contain sensitive personal information or offensive language. Detailed statistics of the SGD and DART datasets can be found in \citet{SGD-dataset-rastogi2020} and \citet{DART_dataset_paper}, respectively. We note that the model may hallucinates information or generates language that appears offensive. Some linguistic phenomena of our DA-T5 models are in Appendix \ref{app:qualitative_study}. It is vital that developers test DA-T5 fully before deployment.

All software packages that our code built on are used as their original intention. Our code is released under the MIT license.

\bibliography{anthology,custom}

\begin{thebibliography}{19}
\expandafter\ifx\csname natexlab\endcsname\relax\def\natexlab#1{#1}\fi

\bibitem[{Chen et~al.(2023)Chen, Lin, and Byrne}]{SchemeGuidedSemanticAccuracy_Chen_2023}
Jinghong Chen, Weizhe Lin, and Bill Byrne. 2023.
\newblock \href {https://doi.org/10.48550/arXiv.2301.12568} {Schema-guided semantic accuracy: Faithfulness in task-oriented dialogue response generation}.
\newblock \emph{CoRR}, abs/2301.12568.

\bibitem[{Gorman(2016)}]{pynini_2016}
Kyle Gorman. 2016.
\newblock \href {https://doi.org/10.18653/v1/W16-2409} {{P}ynini: A {P}ython library for weighted finite-state grammar compilation}.
\newblock In \emph{Proceedings of the {SIGFSM} Workshop on Statistical {NLP} and Weighted Automata}, pages 75--80, Berlin, Germany. Association for Computational Linguistics.

\bibitem[{Hu et~al.(2019)Hu, Khayrallah, Culkin, Xia, Chen, Post, and Durme}]{VectorizedDBA_2019}
J.~Edward Hu, Huda Khayrallah, Ryan Culkin, Patrick Xia, Tongfei Chen, Matt Post, and Benjamin~Van Durme. 2019.
\newblock \href {https://doi.org/10.18653/v1/n19-1090} {Improved lexically constrained decoding for translation and monolingual rewriting}.
\newblock In \emph{Proceedings of the 2019 Conference of the North American Chapter of the Association for Computational Linguistics: Human Language Technologies, {NAACL-HLT} 2019, Minneapolis, MN, USA, June 2-7, 2019, Volume 1 (Long and Short Papers)}, pages 839--850. Association for Computational Linguistics.

\bibitem[{Huang et~al.(2023)Huang, Ke, and Huang}]{PreDAT_pretraining_da_transformer_huang_2023}
Fei Huang, Pei Ke, and Minlie Huang. 2023.
\newblock \href {https://doi.org/10.48550/arXiv.2304.11791} {Directed acyclic transformer pre-training for high-quality non-autoregressive text generation}.
\newblock \emph{CoRR}, abs/2304.11791.

\bibitem[{Huang et~al.(2022)Huang, Zhou, Liu, Li, and Huang}]{DA-Transformer-pmlr-v162-huang22m}
Fei Huang, Hao Zhou, Yang Liu, Hang Li, and Minlie Huang. 2022.
\newblock \href {https://proceedings.mlr.press/v162/huang22m.html} {Directed acyclic transformer for non-autoregressive machine translation}.
\newblock In \emph{Proceedings of the 39th International Conference on Machine Learning}, volume 162 of \emph{Proceedings of Machine Learning Research}, pages 9410--9428. PMLR.

\bibitem[{IEEE(2004)}]{RegularExpressionIEEEStandard}
The Open~Group IEEE. 2004.
\newblock \emph{Chapter 9: Regular Expressions}, ieee std 1003.1, 2004 edition edition, volume~6, chapter~9. IEEE.
\newblock Archived from the original on 2011-12-02. Retrieved 2011-12-13.

\bibitem[{Kale and Rastogi(2020)}]{T2G2_TODS_Kale_and_Rastogi}
Mihir Kale and Abhinav Rastogi. 2020.
\newblock \href {https://doi.org/10.18653/v1/2020.emnlp-main.527} {Template guided text generation for task-oriented dialogue}.
\newblock In \emph{Proceedings of the 2020 Conference on Empirical Methods in Natural Language Processing, {EMNLP} 2020, Online, November 16-20, 2020}, pages 6505--6520. Association for Computational Linguistics.

\bibitem[{Li et~al.(2020)Li, Ding, Liu, Hu, and Durme}]{DisjuctivePositiveConstraints(DPC)_2020}
Zhongyang Li, Xiao Ding, Ting Liu, J.~Edward Hu, and Benjamin~Van Durme. 2020.
\newblock \href {https://doi.org/10.24963/ijcai.2020/502} {Guided generation of cause and effect}.
\newblock In \emph{Proceedings of the Twenty-Ninth International Joint Conference on Artificial Intelligence, {IJCAI} 2020}, pages 3629--3636. ijcai.org.

\bibitem[{Ma et~al.(2023)Ma, Shao, Gui, Zhang, and Feng}]{FuzzyAlignmentsInDATransformerMa2023}
Zhengrui Ma, Chenze Shao, Shangtong Gui, Min Zhang, and Yang Feng. 2023.
\newblock \href {https://openreview.net/pdf?id=LSz-gQyd0zE} {Fuzzy alignments in directed acyclic graph for non-autoregressive machine translation}.
\newblock In \emph{The Eleventh International Conference on Learning Representations, {ICLR} 2023, Kigali, Rwanda, May 1-5, 2023}. OpenReview.net.

\bibitem[{Mohri et~al.(2002)Mohri, Pereira, and Riley}]{MohriWFST_SpeechRecognition_2002}
Mehryar Mohri, Fernando Pereira, and Michael Riley. 2002.
\newblock \href {https://doi.org/10.1006/csla.2001.0184} {Weighted finite-state transducers in speech recognition}.
\newblock \emph{Comput. Speech Lang.}, 16(1):69--88.

\bibitem[{Nan et~al.(2021)Nan, Radev, Zhang, Rau, Sivaprasad, Hsieh, Tang, Vyas, Verma, Krishna, Liu, Irwanto, Pan, Rahman, Zaidi, Mutuma, Tarabar, Gupta, Yu, Tan, Lin, Xiong, Socher, and Rajani}]{DART_dataset_paper}
Linyong Nan, Dragomir~R. Radev, Rui Zhang, Amrit Rau, Abhinand Sivaprasad, Chiachun Hsieh, Xiangru Tang, Aadit Vyas, Neha Verma, Pranav Krishna, Yangxiaokang Liu, Nadia Irwanto, Jessica Pan, Faiaz Rahman, Ahmad Zaidi, Mutethia Mutuma, Yasin Tarabar, Ankit Gupta, Tao Yu, Yi~Chern Tan, Xi~Victoria Lin, Caiming Xiong, Richard Socher, and Nazneen~Fatema Rajani. 2021.
\newblock \href {https://doi.org/10.18653/v1/2021.naacl-main.37} {{DART:} open-domain structured data record to text generation}.
\newblock In \emph{Proceedings of the 2021 Conference of the North American Chapter of the Association for Computational Linguistics: Human Language Technologies, {NAACL-HLT} 2021, Online, June 6-11, 2021}, pages 432--447. Association for Computational Linguistics.

\bibitem[{Papineni et~al.(2002)Papineni, Roukos, Ward, and Zhu}]{BLEU_metric_paper}
Kishore Papineni, Salim Roukos, Todd Ward, and Wei-Jing Zhu. 2002.
\newblock \href {https://doi.org/10.3115/1073083.1073135} {Bleu: A method for automatic evaluation of machine translation}.
\newblock In \emph{Proceedings of the 40th Annual Meeting on Association for Computational Linguistics}, ACL '02, page 311–318, USA. Association for Computational Linguistics.

\bibitem[{Post and Vilar(2018)}]{DynamicBeamAllocation(DBA)_2018}
Matt Post and David Vilar. 2018.
\newblock \href {https://doi.org/10.18653/v1/n18-1119} {Fast lexically constrained decoding with dynamic beam allocation for neural machine translation}.
\newblock In \emph{Proceedings of the 2018 Conference of the North American Chapter of the Association for Computational Linguistics: Human Language Technologies, {NAACL-HLT} 2018, New Orleans, Louisiana, USA, June 1-6, 2018, Volume 1 (Long Papers)}, pages 1314--1324. Association for Computational Linguistics.

\bibitem[{Qian et~al.(2021)Qian, Zhou, Bao, Wang, Qiu, Zhang, Yu, and Li}]{GlancingTransformer_GLAT_2021}
Lihua Qian, Hao Zhou, Yu~Bao, Mingxuan Wang, Lin Qiu, Weinan Zhang, Yong Yu, and Lei Li. 2021.
\newblock \href {https://doi.org/10.18653/v1/2021.acl-long.155} {Glancing transformer for non-autoregressive neural machine translation}.
\newblock In \emph{Proceedings of the 59th Annual Meeting of the Association for Computational Linguistics and the 11th International Joint Conference on Natural Language Processing, {ACL/IJCNLP} 2021, (Volume 1: Long Papers), Virtual Event, August 1-6, 2021}, pages 1993--2003. Association for Computational Linguistics.

\bibitem[{Rastogi et~al.(2020)Rastogi, Zang, Sunkara, Gupta, and Khaitan}]{SGD-dataset-rastogi2020}
Abhinav Rastogi, Xiaoxue Zang, Srinivas Sunkara, Raghav Gupta, and Pranav Khaitan. 2020.
\newblock Towards scalable multi-domain conversational agents: The schema-guided dialogue dataset.
\newblock In \emph{Proceedings of the AAAI Conference on Artificial Intelligence}, volume~34, pages 8689--8696.

\bibitem[{Sellam et~al.(2020)Sellam, Das, and Parikh}]{BLEURT_paper}
Thibault Sellam, Dipanjan Das, and Ankur~P. Parikh. 2020.
\newblock \href {https://doi.org/10.18653/v1/2020.acl-main.704} {{BLEURT:} learning robust metrics for text generation}.
\newblock In \emph{Proceedings of the 58th Annual Meeting of the Association for Computational Linguistics, {ACL} 2020, Online, July 5-10, 2020}, pages 7881--7892. Association for Computational Linguistics.

\bibitem[{Shao et~al.(2022)Shao, Ma, and Feng}]{Viterbi_decoding_for_Da-Transformer}
Chenze Shao, Zhengrui Ma, and Yang Feng. 2022.
\newblock \href {https://doi.org/10.18653/V1/2022.FINDINGS-EMNLP.322} {Viterbi decoding of directed acyclic transformer for non-autoregressive machine translation}.
\newblock In \emph{Findings of the Association for Computational Linguistics: {EMNLP} 2022, Abu Dhabi, United Arab Emirates, December 7-11, 2022}, pages 4390--4397. Association for Computational Linguistics.

\bibitem[{{Tyler Barrus}(2018)}]{PySpellchecker}
{Tyler Barrus}. 2018.
\newblock Pyspellchecker: {Pure Python Spell Checking.}
\newblock \url{https://pypi.org/project/pyspellchecker/}.
\newblock Python version: {3}.

\bibitem[{Xiao et~al.(2022)Xiao, Wu, Guo, Li, Zhang, Qin, and Liu}]{NARTextGenerationSurvey_2022}
Yisheng Xiao, Lijun Wu, Junliang Guo, Juntao Li, Min Zhang, Tao Qin, and Tie{-}Yan Liu. 2022.
\newblock \href {https://doi.org/10.48550/arXiv.2204.09269} {A survey on non-autoregressive generation for neural machine translation and beyond}.
\newblock \emph{CoRR}, abs/2204.09269.

\end{thebibliography}

\appendix

\section{Experiment setup details}
\label{app:experiment_setup_details}

\paragraph{Metrics details} For BLEURT, we use the BLEURT-20 checkpoint. For BLEU, we use the \texttt{sacrebleu} implementation. Decoding times are average of three runs on a single A100 GPU for the SGD dataset and on a single V100 GPU for the DART dataset.


\paragraph{Vocabulary for neologism evaluation} From the entire corpus, we extract all space-delimited words, strip punctuation and numbers, and maintain true cases. All words in the test corpus are also added to the evaluation vocabulary without pre-processing. Note that they are not added to the constraint vocabulary for VC decoding to avoid leakage. For the SGD, we also add all words in the slot names, slot values, and slot descriptions from the schema, resulting in a vocabulary of 19,126 words. In evaluation, we only strip punctuation from words in the generated texts. We also use the \texttt{pyspellchecker} library~\cite{PySpellchecker} to check that the word in question is indeed OOV. 


\paragraph{Exact Occurrence Error} We go through the training data to identify subjects/objects that are always present in the ground-truth text. For example, we find that the subject of the relation \texttt{priceRange} always appear in the ground-truth text. Whenever \texttt{priceRange} appears during testing, we treat the string value of its subject as hard lexical constraints. If the string cannot be found in the generated text, an exact occurrence error is flagged. 

\paragraph{Data Preprocessing} We linearize the input dialogue actions or triplets to strings as input to our DA-T5 model. On the SGD, we follow the Schema Guided Linearization by \citet{T2G2_TODS_Kale_and_Rastogi} to process our input data. 
On DART, we process the triplets into arrays of \texttt{``<h> SUBJECT <r> RELATION <t> OBJECT''} where \texttt{<h>}, \texttt{<r>}, and  \texttt{<t>} are special tokens. 

\paragraph{Training hyper-parameters} The DAG vertex size $L$ is determined by the upsample factor $\lambda$ ($L = \lambda \times N$ where $N$ is the input length) with $\lambda=5$ for both the SGD and the DART datasets. 
We use the \texttt{T5-small} architecture with randomly initialized weights to generate vertex embeddings (79.3M trainable parameters).
 We train the model with a learning rate of 1e-4, a batch size of 8 using the AdamW optimizer. Glancing training is used to facilitate training with a constant annealing factor $\tau=1.0$. SGD training took around 13 hours (25 minutes per epoch) on a single A100 GPU including all validation runs. DART training took 24 hours on a single V100 GPU. We find that glancing training is critical to successful training. Without it the model performs poorly (4.6 BLEU on the SGD when decoded with Greedy). 

\paragraph{Target length predictor} Let $x$ be the input length in tokens, $L_{tgt} = \lceil 26.1x+0.4 \rceil$ for the SGD and  $L_{tgt} = \lceil 0.5x+11.9 \rceil$ for DART.  Coefficients are fitted on the validation set. 
We use strictness $A=1$ in LC decoding. 

\paragraph{Beam search} Auto-regressive Beam Search (BS) and Constrained Beam Search (CBS) use beam size $=5$. CBS-DAG uses a base beam size of $4$ with dynamic adjustment (Sec.\ref{app:CBS-DAG_details}). 

\section{Algorithmic details}
\label{app:all_algo_details}

\subsection{DAG-to-WFSA conversion}
\label{app:dag-to-wfsa_algo_detail}

A Weighted FSA (WFSA) consists of states and weighted directed arcs connecting the states.  The outputs (tokens) are labeled on the arcs. DAG-to-WFSA is simply Moore Machine to Mealy Machine conversion by treating DAG vertices as WFSA states and exploding the output tokens at DAG vertices to WFSA arc labels. WFSA arc weights are the sum of negative log-likelihood for state transition and token emission. The best path has maximal likelihood. 

We prune the DAG before conversion to reduce the number of WFSA arcs. For each vertex $u$ in the DAG, we only keep the top $K_e$ tokens and top $K_t$ transitions in descending probabilities. We also keep tokens that appear in the constraint phrases, ensuring there exists paths that realize lexical constraints in the WFSA (Algo.\ref{algo:DAG-to-WFSA(ForceEmit function)}). Algo.\ref{algo:DAG-to-WFSA(naive)} shows pseudo-code.  $\times$ denotes Cartesian product.




\begin{algorithm}[h!]
\caption{DAG to WFSA conversion}
\begin{algorithmic}[1]
\Statex\textbf{Inputs:} DAG vertices $V$, transition matrix $E$, emission matrix $P$, emission degree $K_e$ and transition degree $K_t$. Lexical constraint phrases $\mathcal{C}=[C_1,...,C_M]$.
\State{$\mathcal{E} \gets \emptyset$} 
\For {$u \in \text{topological\_sort}(V)$} 
\State $\mathcal{T}[u] \gets \arg\text{topk}(P[u, :], K_e)$ 
\State $\mathcal{S}[u] \gets \arg\text{topk}(E[u, :], K_t)$ 
\State $\mathcal{T}[u] \gets \mathcal{T}[u] \ \cup $ \Call{ForceEmit}{$u, \mathcal{C}$} 
\Statex\Comment{Forced emission (Algo.\ref{algo:DAG-to-WFSA(ForceEmit function)}})
\For {$t, v \in \mathcal{T}[u] \times \mathcal{S}[u] $} 
\State $w=-(\log P[u, t]+\log E[u, v])$
\State $e \gets \left(u, t, w, v\right)$ 
\State $\mathcal{E} \gets \mathcal{E} \cup \{e\}$
\EndFor
\EndFor
\State Construct the WFSA with edge set $\mathcal{E}$
\end{algorithmic}
\label{algo:DAG-to-WFSA(naive)}
\end{algorithm}

Finding the shortest path has linear complexity in the number of edges because our WFSA is acyclic. The pruning parameters, $K_t$ and $K_e$, trades of completeness with decoding speed. Larger values lead to a more complete WFSA at the cost of longer decoding time. 

\begin{algorithm}[h!]
\caption{The ForceEmit function}
\begin{algorithmic}[1]

\Statex\textbf{Inputs:} Vertex predecessors under top-K transition pruning $N_{K_t}^{-}(v)$. Lexical constraint phrases $\mathcal{C}=[C_1, ..., C_M]$. Emission tokens at all predecessor vertices $\mathcal{T}[\cdot]$
\Function{ForceEmit}{$u, \mathcal{C}$}
    \State $\mathcal{F} \gets \emptyset $
    \For{phrase $C_i \in \mathcal{C}$}
    \For{token $t_j$ in $C_i[:-1]$}
    \For{$v \in N_{K_t}^{-}(u)$}
    \If{$t_j \in \mathcal{T}[v]$}
    \State $\mathcal{F} \gets \mathcal{F} \cup \{t_{j+1}\}$ 
    \Statex\Comment{Force-emit the next token $t_{j+1}$ in phrase $C_i$}
    \EndIf
    \EndFor
    \EndFor
    \EndFor
    \State\Return $\mathcal{F}$
\EndFunction
\end{algorithmic}
\label{algo:DAG-to-WFSA(ForceEmit function)}
\end{algorithm}

\subsection{Vocabulary Constraint}
\label{app:vc_algo_detail}

We elaborate on how to construct the FSAs for vocabulary constraints below:

\paragraph{Dictionary FSA} From the training corpus, we extract space-delimited unigrams, strip numbers and punctuation, sort them in descending frequency, and cutoff at 90\% cumulative frequency. This results in a vocabulary $V$ of 1129 words on the SGD dataset. We then tokenize each unigram with the T5 tokenizer, build FSA that accepts and only accepts the tokenized sequence (e.g. \texttt{``photosynthesis''} $\rightarrow$ \texttt{``\_photo'', ``synthesis''}), and union these FSAs to form the dictionary FSA $A_{dict}$. 

\paragraph{Special token FSA} $A_{spec}$ accepts and only accepts punctuation ``\texttt{\$\&'()*+,-./:;=>?@[]\_}'', start-of-sentence \texttt{<s>}, end-of-sentence token \texttt{</s>}, and T5 tokenizer's start-of-word mark (u2581 ``\texttt{\_}''). 
    
\paragraph{Dynamic FSA}: $A_{dyn}$ is built for each input. Given the entity names, we tokenize them, build FSAs that accepts and only accepts the token sequence for each entity, and take the union. Note that entity names may include space. For example, $A_{dyn}$ may accept ``Hong Kong'' but not the constituent unigrams ``Hong'' and ``Kong''.

\subsection{Length Constraint}
\label{app:lc_algo_detail}

Algo.\ref{algo:DFS-Viterbi} lists the DFS-Viterbi algorithm and the symbol definitions. The recursive relation is given in Eq.\ref{eq:recursive relation for DFS memo}. For each vertex, we memoize the current best string of each length and their costs. The shortest path is recovered with parent pointers.

\begin{equation}
    \delta(u, l+1) = \min_{v \in N^{+}_p(u)} w(u,v) + \delta(v,l)
    \label{eq:recursive relation for DFS memo}
\end{equation}

We fit a first-order linear model to predict target length $L_{tgt}$ from input length. Length is measured in tokens and coefficients are given in Appendix \ref{app:experiment_setup_details}. Enforcing a strict length constraint can lead to incomplete sentences. Therefore, we find the best $l-$length string for $l=1,\ldots,L_{upper}$, where $L_{upper}=\min(L_{tgt}+5, L_{tgt}\times 1.5)$ and introduce an exponential length penalty (Eq.\ref{eq:target length penalty for dfs memo}) similar to BLEU. The candidate with the lowest overall cost $C'$ (Eq.\ref{eq:modified cost with target length penalty for dfs memo}) is chosen as the final generation. We use simple linear regression to specify the length target $L_{tgt}$. 

\begin{gather}
    LP = 
    \begin{cases} 
        \exp\big(A(L_{tgt}/l-1)\big), & \text{if } l < L_{tgt} \\
        1, & \text{otherwise}
    \end{cases} \label{eq:target length penalty for dfs memo} \\
    C' = LP \times \delta(u_s, l)
    \label{eq:modified cost with target length penalty for dfs memo}
\end{gather}

The WFSA software implementation, \texttt{pynini}~\cite{pynini_2016},  allows us to efficiently traverse the WFSA as graphs. Prior to running DFS-Viterbi, we sort the WFSA states topologically and perform epsilon-removal~\cite{MohriWFST_SpeechRecognition_2002}. Epsilon transitions do not have actual token labels, and are removed to prevent over-counting the output length. The WFSA can be topologically sorted because intersection preserves the acyclic property of its input: any cycles will result in strings of unbounded length which cannot be accepted by the acyclic WFSA. 

Let $|V|$ be the number of WFSA states. The space complexity of memoization is $O(L_{tgt}\times |V|)$. The worst-case time complexity is exponential $O(L_{tgt}^{|V|})$. However, we observe a linear time complexity of $O(L_{tgt})$ when applying DFS-Viterbi to our trained DA-T5 model. We attribute the efficiency to: (1) memoization; (2) transition probabilities are concentrated on a few successors. We find that the number of out-going edges after pruning, $|N_{p}^+(u)|$, approximates 1 when $p=0.7$, leading to very efficient search.

\begin{algorithm}
\caption{DFS-Viterbi finds the shortest path with exactly $L_{tgt}$ edges.}
\begin{algorithmic}[1]
\Function{DFS-Viterbi}{$u$, $l$, $\delta$, $L_{tgt}$, $N^{+}$, $w$}
    \State \textbf{Arguments:}
    \State \(u\): current vertex.
    \State \(l\): target length (number of edges) from vertex \(u\) to a final vertex.
    \State \(\delta\): memoization table storing shortest distance to vertex \(u\) with exactly \(l\) edges.
    \State \(F\): set of final states (vertices).
    \State \(N^{+}_p(u)\): minimal set of successors of vertex \(u\) with cumulative probability $>p$.
    \State \(w(u, v)\): edge weight from vertex \(u\) to \(v\).
    \If{$v$ is in $F$}
        \State \Return $0$
    \EndIf
    \If{$\delta[u, l]$ is not NULL}
        \State \Return $\delta[u, l]$
    \EndIf
    \State $\text{min\_distance} \gets \infty$
    \ForAll{$v \in N^{+}(u)$}
        \State $\text{dist} \gets w(u,v) +$ \Call{DFS-Viterbi}{$v, l+1, \delta, F, N^{+}, w$}
        \If{$\text{dist} < \text{min\_distance}$}
            \State $\text{min\_distance} \gets \text{dist}$
        \EndIf
    \EndFor
    \State $\delta[u, l] \gets \text{min\_distance}$
    \State \Return $\text{min\_distance}$
\EndFunction
\end{algorithmic}
\label{algo:DFS-Viterbi}
\end{algorithm}

\subsection{Constrained Beam Search for Directed Acyclic Graphs (CBS-DAG)}
\label{app:CBS-DAG_details}

CBS-DAG follows the beam expansion and pruning rules in Dynamic Beam Allocation (DBA)~\cite{DynamicBeamAllocation(DBA)_2018}. Let $K$ be the beam size. At each vertex transition, CBS-DAG extends the beam with the top-$K$ tokens from model prediction, the next token in active constraints, and the first token in non-active constraints. Active constraints are identified by the KMP string-matching algorithm. 
After beam expansion, we regroup the candidates into ``banks'' by the number of unmet constraint tokens and retain the most likely candidate within each bank. We dynamically adjust the beam size such that beam size is always larger than the number of non-empty banks (i.e., the number of constraint tokens plus one). 



\section{Further Analysis}
\label{app:further_analysis}

\paragraph{DA-T5 produces sparse DAGs} We find that DA-T5 learns to produce a sparse DAG in the following sense: on average, each vertex has 1.68 transitions with probability $>0.2$ and 1.58 emissions with probability $>0.2$  after training. These statistics are computed over the validation set, and explain why we can prune aggressively during WFSA-to-DAG conversion (top-3 for the SGD and top-5 for DART) for speed without much loss of information.

\section{Qualitative Study}
\label{app:qualitative_study}

\begin{figure*}[h!]
    \centering
    \includegraphics[width=1.0\textwidth]{qualitative_cases_controldag.pdf}
    \caption{Case study comparing DA-T5 with Control-DAG, Joint Viterbi, and CBS-DAG decoding on the SGD dataset.}
    \label{fig:qualitative_cases}
\end{figure*}

\end{document}